%% file: main.tex
\documentclass[10pt,twocolumn,letterpaper]{article}

 \usepackage{cvpr}              % To produce the CAMERA-READY version
\input{preamble}
\definecolor{cvprblue}{rgb}{0.21,0.49,0.74}
\usepackage[pagebackref,breaklinks,colorlinks,allcolors=cvprblue]{hyperref}
\usepackage[table]{xcolor}
\usepackage{tcolorbox}
\tcbuselibrary{skins, breakable}
\usepackage{fvextra}
\DefineVerbatimEnvironment{PromptJSON}{Verbatim}{
    breaklines=true,
    breaksymbolleft={},
    breakanywhere=true
}

\DefineVerbatimEnvironment{PromptTEXT}{Verbatim}{
    breaklines=true,
    breakanywhere=true,
    breaksymbolleft={},
    fontsize=\small
}

\def\paperID{2124} % *** Enter the Paper ID here
\def\confName{CVPR}
\def\confYear{2026}

\title{CapNav: Benchmarking Vision Language Models on Capability-conditioned Indoor Navigation}

\author{Xia Su\thanks{Both authors contributed equally to this research.}\\
University of Washington\\
Seattle, WA, USA\\
{\tt\small xiasu@cs.washington.edu}
\and
Ruiqi Chen\footnotemark[1]\\
University of Washington\\
Seattle, WA, USA\\
{\tt\small ruiqich@uw.edu}
\and
Benlin Liu\\
University of Washington\\
Seattle, WA, USA\\
{\tt\small liubl@cs.washington.edu}
\and
Jingwei Ma\\
University of Washington\\
Seattle, WA, USA\\
{\tt\small jingweim@cs.washington.edu}
\and
Zonglin Di\\
University of California, Santa Cruz\\
Santa Cruz, CA, USA\\
{\tt\small zdi@ucsc.edu}
\and
Ranjay Krishna\\
University of Washington\\
Seattle, WA, USA\\
{\tt\small ranjay@cs.washington.edu}
\and
Jon Froehlich\\
University of Washington\\
Seattle, WA, USA\\
{\tt\small jonf@cs.washington.edu}
}

\begin{document}
\twocolumn[{%
\renewcommand\twocolumn[1][]{#1}%
\maketitle

\begin{center}
    \includegraphics[width=\linewidth]{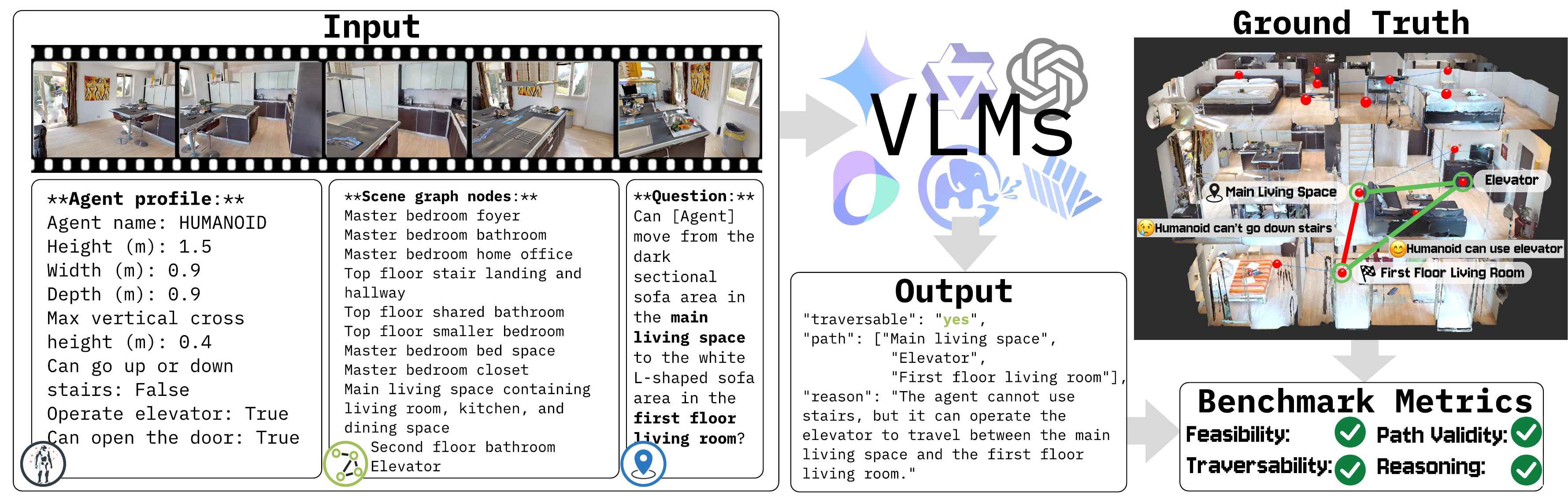}
    \captionof{figure}{We introduce Capability-Conditioned Navigation (\textit{\textbf{CapNav}}), a benchmark designed to evaluate how well VLMs can navigate complex indoor spaces given an agent’s specific physical and operational capabilities. CapNav inputs (1) a tour video of an indoor space, (2) nodes of its navigation graph, (3) an agent's mobility profile, and (4) a navigation task, and evaluates VLM outputs in task feasibility, path validity, route traversability, and reasoning validity.}
    \label{fig:teaser}
\end{center}

\vspace{2em}
}]

\input{sec/0_abstract}    
\input{sec/1_intro}
\input{sec/2_related_works_new}

\input{sec/3_benchmark}

\input{sec/4_experiments_new}

\input{sec/5_findings_polished}
\input{sec/6_conclusion}
{
    \small
    \bibliographystyle{ieeenat_fullname}
    \bibliography{main}
}

\input{sec/X_suppl}

\end{document}

%% file: sec/0_abstract.tex
\begin{abstract}
Vision-Language Models (VLMs) have shown remarkable progress in Vision-Language Navigation (VLN), offering new possibilities for navigation decision-making that could benefit both robotic platforms and human users. However, real-world navigation is inherently conditioned by the agent’s mobility constraints. For example, a sweeping robot cannot traverse stairs, while a quadruped can. We introduce Capability-Conditioned Navigation (\textit{\textbf{CapNav}}), a benchmark designed to evaluate how well VLMs can navigate complex indoor spaces given an agent’s specific physical and operational capabilities. CapNav defines five representative human and robot agents, each described with physical dimensions, mobility capabilities, and environmental interaction abilities. CapNav provides 45 real-world indoor scenes, 473 navigation tasks, and 2365 QA pairs to test if VLMs can traverse indoor environments based on agent capabilities. We evaluate 13 modern VLMs and find that current VLM's navigation performance drops sharply as mobility constraints tighten, and that even state-of-the-art models struggle with obstacle types that require reasoning on spatial dimensions. We conclude by discussing the implications for capability-aware navigation and the opportunities for advancing embodied spatial reasoning in future VLMs. The benchmark is available at \href{https://github.com/makeabilitylab/CapNav}{https://github.com/makeabilitylab/CapNav}

%Old version
%Vision Language Models (VLM) have demonstrated promising performance in spatial reasoning tasks, including vision language navigation (VLN), opening pathways towards plug-and-use navigation decision making that can benefit both robotic platforms and human users. However, real-world navigation tasks are highly conditioned by both environmental factors and the agent's mobility constraints. For example, never navigate a wheeled robot to go up stairs. Yet these mobility constraints are never discussed in VLM's application in VLN.
%We introduce Capability-conditioned Navigation (CapNav), a benchmark that examines how well can VLMs navigate complex indoor spaces based on an agent’s capabilities. CapNav provides detailed descriptions of the agent's physical footprint, vertical mobility, and environmental interaction capabilities, and tests VLMs on whether and how the agent can traverse complex indoor environments. Our dataset builds on 45 real-world indoor scenes, supporting 5 representative mobility types, 473 navigation tasks, and over 5000 manual annotations for traversability. CapNav enables systematic evaluation of VLM's capability-aware spatial reasoning and navigation. Evaluating 10 state-of-the-art VLMs, we reveal significant performance gaps on more constrained mobility settings, highlighting limitations in VLM's spatial understanding and motivating future research on capability-aware navigation.
\end{abstract}

%% file: sec/1_intro.tex
\section{Introduction}

\begin{figure*}
    \centering
    \includegraphics[width=0.9\linewidth]{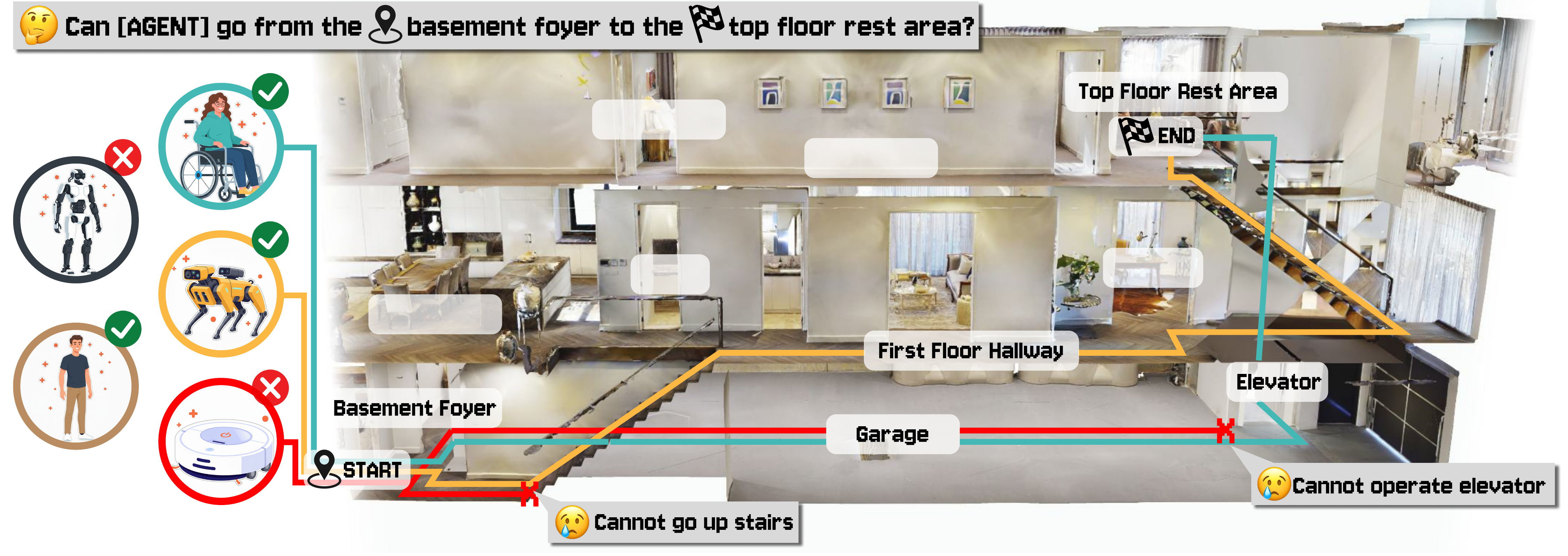}
    \caption{The CapNav benchmark evaluates whether VLMs can correctly ground differences in agent mobility capabilities when generating navigation plans. This example demonstrates a navigation task that has different feasibility and path for different agents.}
    %The CapNav benchmark evaluates how well can VLMs understand the difference of mobility capability and generate capability-conditioned navigation plans.}
    \label{fig:example}
\end{figure*}

%Vision-and-language navigation (VLN) is a task in which agents navigate through environments following natural language instructions ~\cite{anderson2018r2r,zhu2020vision}. In parallel, general-purpose vision–language models (VLMs) and multimodal LLMs have rapidly improved at grounding language in images and video--- NEED SOME MORE SPECIFIC RW EXAMPLES.
%These models are now routinely used as drop‑in navigation brains or planning modules for robots and assistive systems~\cite{shah2023lmnav, zhou2023navgpt, goetting2024end}. Yet we lack a systematic understanding of how well these models generalize across agents with different physical dimensions and mobility constraints.

As vision–language models (VLMs) advance in visual grounding \cite{shi2024aware} and spatial reasoning \cite{cheng2024spatialrgpt}, they are now commonly used as drop-in “navigation assistants” or planning modules for robots and assistive systems~\cite{shah2023lmnav, zhou2023navgpt, goetting2024end}. Guiding people to reach destinations \cite{li2025fine,goetting2024end} and controlling robots in movement decisions \cite{huang2022visual}, they are widely applied in open-ended scenarios where solutions can be pluralistic, and their plausibility depends on the agents' movement capabilities. 

Consider the multi‑story building shown in \autoref{fig:example}. To go from the basement foyer to the top-floor rest area, a wheelchair user should use the elevator, but a quadruped robot can only go up the stairwell (since it cannot operate the elevator). Similarly, a non-disabled human may squeeze through a cluttered corridor, while a wide humanoid robot cannot. Without awareness of capability, VLM may recommend routes that are infeasible or even unsafe for a given agent ~\cite{li2023pope, guan2024hallusionbench}.

%Numerous frameworks \cite{yin2024navigation,goetting2024end,anderson2018r2r, koh2021pathdreamer, hwang2023meta} and benchmarks \cite{wu2025spatial, yang2025thinking, feng2025video, song2025towards} have been proposed to evaluate these navigation capabilities. Most work fail to include complex enough space that reflects real-world navigation obstacles. Some already consider embodiment but not precise enough or generalize enough. However, existing evaluation targets XXXX,XXX, and XXX. Thus, they fail to capture the pluralistic nature and capability-conditioned validity of navigation tasks.

%Such mobility constraints and pluralistic nature have been neglected in past vision language navigation research. Existing frameworks and benchmarks target embodiment-agnostic goal-reaching in simulated environments \cite{anderson2018r2r, krantz2020vlnce, wang2025rethinking, song2025towards} , or as simplified navigational VQA \cite{feng2025video, wu2025spatial, yang2025thinking, xu2025spatialbench}. They also evaluate trajectory fidelity compared to a single ground truth path \cite{jain2019stay, ilharco2019ndtw, anderson2018eval} instead of multiple possible paths. They fail to capture the pluralistic nature and capability-conditioned validity of navigation tasks, which are crucial for the growing needs of embodied controls.

Such mobility constraints and the pluralistic nature of navigation remain underexplored in prior vision–language navigation research. Existing frameworks and benchmarks evaluate embodiment-agnostic goal-reaching in simulated or graph-based environments~\cite{anderson2018r2r, krantz2020vlnce, wang2025rethinking, song2025towards}, or reduce navigation to simplified VQA-style spatial reasoning tasks~\cite{feng2025video, wu2025spatial, yang2025thinking, xu2025spatialbench}. Moreover, evaluation protocols typically measure trajectory fidelity against a single annotated ground-truth path~\cite{jain2019stay, ilharco2019ndtw, anderson2018eval}. Consequently, these paradigms overlook the capability-conditioned validity and solution multiplicity that characterize real-world navigation — limitations that become increasingly critical as VLMs are deployed in embodied control and assistive scenarios.

%In this case, a single navigation instruction (“go to the master bedroom”) should allow for different trajectories conditioned on the agent’s capability profile. It remains unclear how VLMs behave across different agents when facing mobility obstacles such as stairs, narrow doors, and cluttered passageways.

%Existing VLN datasets and benchmarks—R2R~\cite{anderson2018r2r}, RxR~\cite{ku2020rxr}, REVERIE~\cite{qi2020reverie}, and VLN‑CE~\cite{krantz2020vlnce}—have been pivotal for progress in this task, but  Even recent LLM‑enhanced systems~\cite{shah2023lmnav, zhou2023navgpt} and large‑scale 3D scene resources like MP3D and HM3D~\cite{chang2017matterport3d, ramakrishnan2021hm3d} do not isolate \emph{capability‑conditioned} performance. Consequently, it remains unclear how the same VLM behaves across different embodiments when facing mobility obstacles such as stairs, narrow doors, and cluttered passageways.

We introduce \textit{CapNav}, a benchmark for VLM's navigation validity across varying mobility constraints. We consider five representative embodiments: adults with no motor disabilities, wheelchair users, sweeping robots, humanoid robots, and quadrupedal robots. We use real-world 3D scans \cite{chang2017matterport3d, ramakrishnan2021hm3d} and annotate navigation graphs ~\cite{anderson2018r2r, savva2019habitat, krantz2020vlnce} to allow plural solutions and per-edge traversability analysis. CapNav inputs the agent description, realistic indoor videos, key spatial reference nodes, and a navigation task as``from A to B". We then examine VLM outputs for the prediction of feasibility, the navigation path, and reasoning validity. In total, CapNav contains 45 scenes, 2365 navigation tasks, and 5075 traversability annotations as ground truth. 

%Pronounce the findings!
Using the CapNav benchmark, we systematically study 13 opensource and proprietary VLMs under varying inference settings. Through quantitative and qualitative analysis, we diagnose three major observations: (1) \textbf{mobility degradation}: navigation performance drops significantly when considering mobility capabilities; (2) \textbf{visual bottleneck}: increasing visual input does not always lead to a performance boost in CapNav; (3) \textbf{spatial dimension blindness}: all models fail to detect non-traversability arising from insufficient spatial clearance (e.g., narrow passages or limited turning radius).

%discussion and reflection from the results.
%These results indicate that more work needs to be done in VLM-based navigation to ensure solid output for specific embodiments, especially in high-stake scenarios. Also, current VLM architectures and training schemes should be improved to really enable spatial reasoning in complex environments, especially those that require many visual frames; Also, models' training should be improved to better address mobility limitations and spatial dimension reasoning.

These results highlight that substantial progress is still needed before VLM-based navigation can reliably support embodiment-specific deployment, particularly in safety-critical scenarios. Moreover, current VLM architectures and training paradigms remain limited in their ability to integrate spatial information across multiple visual frames and to perform precise metric reasoning in complex environments. Advancing embodied navigation, therefore, requires improvements in large-scale visual integration, geometry-grounded reasoning, and training objectives that explicitly encode mobility constraints and spatial dimension awareness.

Our contributions are threefold:
\begin{enumerate}
    \item \textbf{The CapNav benchmark.} We introduce \textsc{CapNav}, a capability-conditioned VLN benchmark that evaluates how VLMs navigate complex indoor environments under realistic mobility constraints across five representative human and robot embodiments.
    \item \textbf{Comprehensive evaluation of VLMs.} We conduct a head-to-head assessment of 13 state-of-the-art VLMs on capability-aware feasibility prediction, path validity, route traversability, and reasoning quality.  
    \item \textbf{Guidelines and resources.} We analyze the effects of input frame rates, model settings, obstacle types, and failure modes, providing actionable guidelines for capability-aware navigation. We release the full dataset—videos, tasks, agent profiles, and 5k+ traversability annotations—along with an interactive annotation interface for extending CapNav to new agents and environments.
\end{enumerate}

%% file: sec/2_related_works_new.tex
\section{Related Work}

\subsection{Vision Language Navigation and Benchmarks}
\label{subsec:vln}
Vision-Language Navigation (VLN) studies how an agent follows natural-language instructions using grounded visual observations to traverse an environment and reach a goal~\cite{anderson2018r2r,zhu2020vision}. Since the introduction of Room-to-Room (R2R)~\cite{anderson2018r2r}, which casts indoor navigation as step-by-step action selection on a sparse connectivity graph, a wide range of datasets and benchmarks have expanded VLN along multiple axes. Indoor instruction-following has been scaled in path length, diversity, and grounding density (e.g., R4R~\cite{jain2019stay} and RxR~\cite{ku2020room}); broader urban/street settings have also been explored (e.g., TOUCHDOWN~\cite{chen2019touchdown}, StreetLearn~\cite{mirowski2019streetlearn}, CityLearn~\cite{chancan2020citylearn}, CityWalker~\cite{liu2025citywalker}, Talk2Nav~\cite{vasudevan2019talk2nav}). Other benchmarks couple navigation with object grounding or task execution (e.g., REVERIE~\cite{qi2020reverie}, ALFRED~\cite{shridhar2020alfred}, ObjectNav~\cite{batra2020objectnav}, EmbodiedBench~\cite{yang2025embodiedbench}), while newer efforts probe long-horizon and open-ended embodied behaviors (e.g., GOAT-bench~\cite{khanna2024goat} and NavBench~\cite{qiao2025navbench}).

Beyond task variants, VLN benchmarks differ in their observation and interaction assumptions. The canonical setting is \emph{interactive}: an agent explores incrementally, receiving egocentric observations and accumulating spatial knowledge step-by-step. In contrast, a \emph{passive} formulation provides global context, such as panoramic scans or pre-recorded videos, aiming to evaluate high-level spatial reasoning while reducing confounds from exploration and low-level control. For example, VideoNavQA~\cite{cangea2019videonavqa} replaces interactive navigation with oracle trajectory videos to isolate embodied visual reasoning; NRNS~\cite{hahn2021no} learns image-goal navigation from passive roaming videos without online RL or a simulator; and PONI~\cite{ramakrishnan2022poni} trains an object-search module interaction-free from offline semantic maps. OpenEQA~\cite{majumdar2024openeqa} further supports an episodic-memory (video) setting that evaluates spatial memory and reasoning over recorded tours and scans. Under such passive formulations, the task increasingly resembles route planning and feasibility assessment; following prior work~\cite{cangea2019videonavqa,majumdar2024openeqa}, we still refer to this setting as navigation.

Our goal---benchmarking VLM navigation validity under explicit mobility constraints---benefits from global observation, since it allows models to reason about traversability and alternative routes across the entire scene while isolating embodiment constraints from exploration and execution noise. We further adopt a graph-based spatial abstraction~\cite{anderson2018r2r,ku2020room} to support pluralistic solutions and enable fine-grained evaluation of path feasibility and validity. This complements existing VLN benchmarks by enabling structured, capability-conditioned comparisons across real-world environments and embodiments.

\subsection{Navigation with mobility constraints}
Past navigation research has modeled a wide range of mobility constraints, including accessibility for wheelchair users~\cite{fleiner2016accessible,sahoo2024autonomous,zhang2024shared}, assistance for people who are blind or low vision~\cite{weiss2019navigationagentsvisuallyimpaired,feghali2024comprehensive,okolo2024assistive}, and robotics embodiments such as quadrupeds~\cite{lee2024learning,liu2024safe,hoeller2024anymal}, humanoids~\cite{chen2025hand}, and vehicles operating over varied terrain~\cite{li2024vln,kabir2025terrain}. Some works also consider settings where multiple mobility profiles co-exist, either through simulator diversity~\cite{wu2024metaurban,wu2025towards,xie2025vid2sim} or by modeling user preferences~\cite{li2025accessibility}. Collectively, these studies highlight that embodiment constraints materially change feasible routes and navigation strategies.

However, most evaluations remain siloed---focusing on a single embodiment, environment family, or constraint type---making cross-embodiment comparison difficult on matched tasks. Recent efforts begin to address this gap: VLN-PE (introduced with a holistic study of physical and visual disparities) enables systematic cross-embodiment evaluation in a physically grounded simulator setting~\cite{wang2025rethinking}; NaviTrace~\cite{windecker2025navitrace} evaluates embodiment-conditioned navigation via 2D trace prediction in real-world images; and VAMOS~\cite{castro2025vamos} introduces an affordance-grounded hierarchical planner to modulate plans by embodiment capabilities. Nonetheless, existing benchmarks either emphasize simulator-centric embodiment effects or cast the problem as trace/plan prediction without explicitly validating pluralistic, capability-conditioned route feasibility in complex indoor spaces with realistic navigation obstacles (e.g., multi-floor transitions, bottlenecks, and accessibility-specific affordances).

%% file: sec/3_benchmark.tex
\section{The CapNav Benchmark}
\label{sec:capnav}

We introduce \emph{Capability-conditioned Navigation} (CapNav), a benchmark for assessing how well can VLMs navigate in built environments given particular physical capabilities. 

\subsection{Problem Statement}
\label{subsec:probstate}
In the CapNav task, each query instance is defined by a \emph{Space--Task--Capability} triple
\[
\langle \mathcal{S}, \tau, \mathbf{a} \rangle,
\]
where the space $\mathcal{S}$ is represented by a touring video and a set of key spatial nodes; $\tau$ is a natural-language navigation goal specifying the source and target locations; and $\mathbf{a}$ is an agent profile encoding physical dimensions and operational abilities.  
Given $\langle \mathcal{S}, \tau, \mathbf{a} \rangle$, a vision language model produces
\[
(\hat{y}, \hat{P}, \hat{\rho}) \;=\; f_\theta(\mathcal{S}, \tau, \mathbf{a}),
\]
where $\hat{y}\!\in\!\{0,1\}$ denotes task feasibility (\textsc{can} / \textsc{cannot} complete), $\hat{P}=[v_0,\dots,v_m]$ is a node sequence representing the proposed navigation path, and $\hat{\rho}$ is a concise rationale explaining infeasibility when $\hat{y}\!=\!0$.  
 
This formulation enables CapNav to holistically evaluate a VLM’s capability-aware navigation ability.  We assess performance from four complementary aspects: (i) \emph{navigation feasibility}, (ii) \emph{path validity}, (iii) \emph{traversability accuracy}, and (iv) \emph{reasoning quality} for infeasible cases.

\begin{figure*}
    \centering
    \includegraphics[width=1\linewidth]{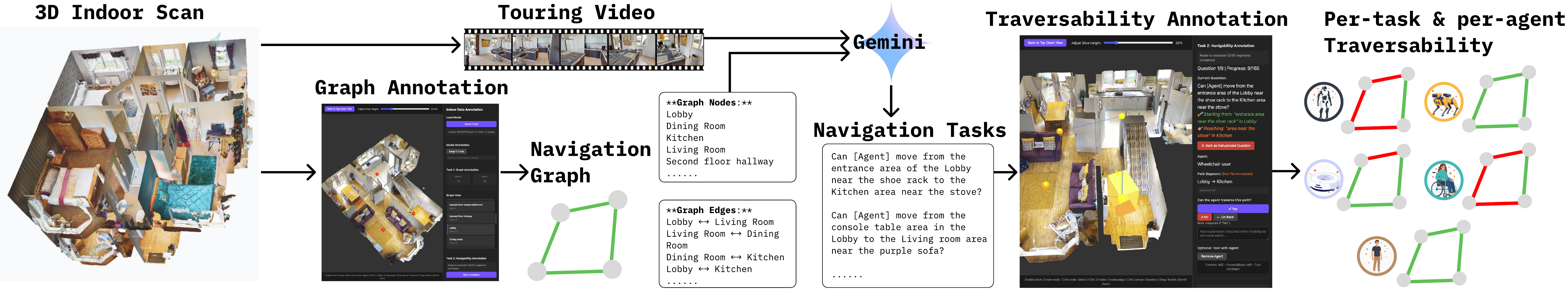}
    \caption{Overview of CapNav's data construction: Starting from a 3D indoor scan, we manually record a touring video and a navigation graph. We then use Gemini to generate natural language navigation tasks. Finally, per-task and per-agent traversability are annotated by manually controlling agents in the annotation interface.}
    \label{fig:data}
\end{figure*}

\paragraph{Space.}
Each space $\mathcal{S}$ is a scanned 3D indoor environment comprising multiple interconnected rooms. As discussed in \autoref{subsec:vln}, we provide spatial information as a rendered touring video $\mathcal{X}=\{x_t\}_{t=1}^{T}$. To better ground the navigation to language description, we manually create a connectivity graph $\mathcal{G}=(\mathcal{V},\mathcal{E})$ over semantically meaningful spatial nodes (e.g., \textit{kitchen}, \textit{entry foyer}, \textit{hallway}) and edges that denote directly walkable connections between the nodes. At inference, we only provide the video and the node list. %Nodes $v\!\in\!\mathcal{V}$ are manually annotated with semantic labels $c(v)$ and approximate 3D positions, while edges $e=(u,v)\!\in\!\mathcal{E}$ denote bidirectional, walkable connections. %The graph is created using an annotation interface that supports node placement, semantic description, and bidirectional edge creation; see Fig.~\ref{fig:annotation-ui} for the UI and example of an annotated graph.

\paragraph{Task.}
A navigation task $\tau$ is a natural-language instruction specifying a movement goal from a source node to a target node (\textit{e.g.,} ``\emph{From the wooden cabinet in the entrance foyer, go to the desk area in the master bedroom.}''). 
Formally, each goal $\tau$ can be expressed as a directed pair:
\[
\tau : (v_{\mathrm{src}}, d_{\mathrm{src}}) \;\rightarrow\; (v_{\mathrm{tgt}}, d_{\mathrm{tgt}}),
\quad v_{\mathrm{src}}, v_{\mathrm{tgt}} \in \mathcal{V},
\]
where $v_{\mathrm{src}}$ and $v_{\mathrm{tgt}}$ denote the source and target nodes in $\mathcal{V}$, and $d_{\mathrm{src}}$ and $d_{\mathrm{tgt}}$ are their corresponding natural-language descriptions specifying the precise starting and ending positions within those nodes (\textit{e.g.,} ``\emph{at the wooden cabinet in the entrance foyer}'', ``\emph{beside the desk area in the master bedroom}'').  
This representation captures both the high-level spatial nodes and the finer-grained textual localization needed for realistic indoor navigation reasoning.

% Formally, each task is associated with a grounded node pair $(v_{\mathrm{src}}, v_{\mathrm{tgt}})$ where $v_{\mathrm{src}}, v_{\mathrm{tgt}}\in\mathcal{V}$.  
% Both the touring video $\mathcal{X}$ and the graph $\mathcal{G}$ are provided to the model. Depending on model constraints, the video input can be supplied as the full $2$\,FPS sequence or as a subset of sampled frames.

\paragraph{Agent Profiles}

We specify the mobility capabilities $\mathbf{a}$ with five distinct but representative agent profiles that cover human mobility and common robot platforms. \textbf{Adult with no motor disabilities} represent a default condition where all indoor routes can be achieved; \textbf{Wheelchair users} cannot use stairs and require clearance for passing and turning; \textbf{Humanoid robot} cannot go up/down stairs and require clearance for passing; \textbf{Sweeping robots} require a flat floor surface; \textbf{Quadrupedal robots} can traverse most indoor spaces but cannot operate doors/elevators.

\begin{figure}
    \centering
    \includegraphics[width=1\linewidth]{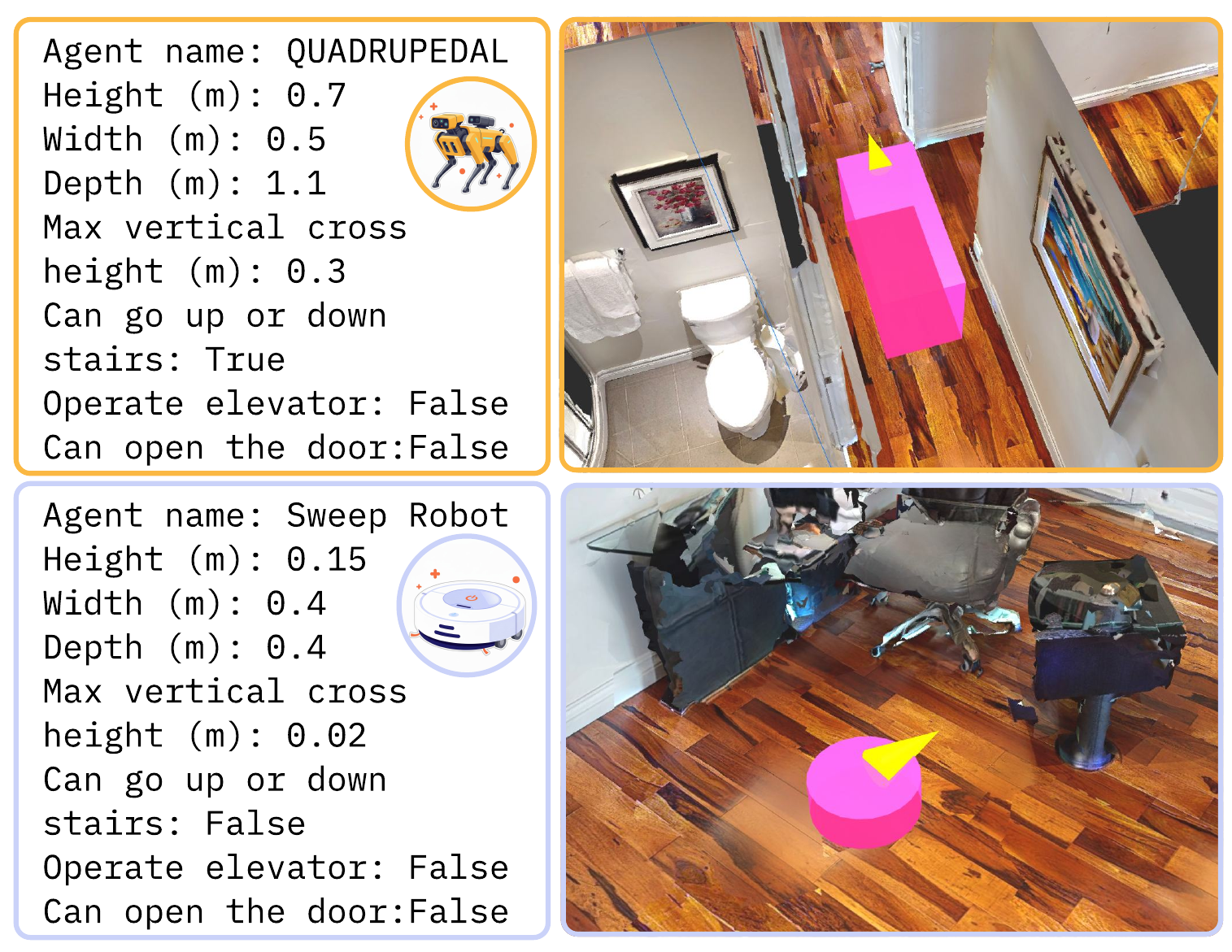}
    \caption{Examples of CapNav's agent profiles. Left: each profile specifies the physical dimensions and functional capabilities. Right: the agents' physical dimensions are rendered and maneuvered in the 3D scenes to help confirm traversability.}
    \label{fig:agents}
\end{figure}

Each profile is described by a capability json
\(\mathbf{a} = (\phi,\kappa,\mu)\),
where $\phi$ captures the physical footprint; $\kappa$ captures vertical traversal limits, including vertical height of a single obstacle, and whether the agent can cross continuous stairs; and $\mu$ captures operation/manipulation abilities. These attributes directly affect traversability on edges $\mathcal{E}$ and on overall tasks $\tau$. See \autoref{fig:agents} for examples. %A full list of capability profiles can be found in Appendix ??.

\subsection{Ground Truth of Traversability}
\label{subsec:gt}

Traversability is manually annotated per-task and per-agent type. They are binary labels at the \emph{edge} level, allowing multiple possible routes and providing more precise reasoning if a route is not traversable. For each navigation task $\tau$ and embodiment $\mathbf{a}$, annotators iterate through each possible simple path that connects \(v_{\mathrm{src}}\) and \(v_{\mathrm{tgt}}\) in \(\mathcal{G}\), and assign
\(g_e^{(\mathbf{a})} \in \{0,1\},\) to each edge $e$ on these paths,
indicating whether $e$ is passable under $\mathbf{a}$ given local spatial geometry and mobility capabilities. Reasons for non-traversability are annotated as text descriptions (\textit{e.g.,} ``\emph{cannot go up/down stairs}''). The annotation UI visualizes 3D colliders matching $\phi$ and supports manually controlled moving/turning to verify clearance and turning space (Fig.~\ref{fig:data}). 

Given per-edge labels, we can then define the feasibility ground truth for a task $(s,g)$ as the existence of at least one simple path (an acyclic sequence of distinct nodes) $P(v_{\mathrm{src}}, v_{\mathrm{tgt}})$ that connects \(v_{\mathrm{src}}\) and \(v_{\mathrm{tgt}}\) in \(\mathcal{G}\), where all edges are traversable:
\[
y^\star = \mathbb{I}\!\big[
  \exists\, P(v_{\mathrm{src}}, v_{\mathrm{tgt}})\!:\!
  \forall (u,v)\!\in\!P,\, g^{(\mathbf{a})}_{(u,v)} = 1
\big],
\]

% \[
% y^\star = \mathbb{I}\!\big[\exists P: v_{\mathrm{src}}\!\rightarrow\!v_{\mathrm{tgt}},\;
% \forall (u,v)\!\in\!P,\, g^{(\mathbf{a})}_{(u,v)} = 1 \big].
% \]

% \[
% y^\star \;=\; \mathbb{1}\!\Big[\,\exists \text{ path } P = [v_0{=}s,\dots,v_m{=}g] \text{ with } \prod_{i=0}^{m-1} g_{(v_i,v_{i+1})}^{(\mathbf{a})} = 1 \,\Big].
% \]
%We compute $y^\star$ via Dijkstra on $\mathcal{G}$ while masking edges with $g_e^{(\mathbf{a})}=0$.

\subsection{Dataset}
\label{subsec:dataset}

We collect 3D indoor scenes from HM3D~\cite{ramakrishnan2021hm3d} and Matterport3D~\cite{chang2017matterport3d}, and record touring videos in the Habitat simulator~\cite{savva2019habitat} by manually navigating each space. Videos are rendered at 2FPS from human-eye height ($1.5m$) with a $75°$ field of view to mimic casual handheld walkthroughs. For each scene, we construct a navigation graph via a custom annotation interface: nodes $v\!\in\!\mathcal{V}$ are annotated with semantic labels $c(v)$ and approximate 3D positions, and edges $e=(u,v)\!\in\!\mathcal{E}$ denote bidirectional, directly walkable connections. We provide each scene’s video and node list to Gemini~2.5~Pro to generate navigation tasks, manually verify task validity, and annotate edge-level traversability for five embodiments, including reasons for non-traversable cases. After filtering out scenes with major holes, disconnected subspaces, repetitive semantics, or trivial layouts, we retain 45 indoor scenes (avg.\ 160.38\,s per video, 13.8 nodes, 14.5 edges), yielding 2{,}365 navigation tasks and 5{,}075 traversability labels (3{,}945 positive, 1{,}130 negative). Feasibility varies substantially across agent types in both task-level and edge-level ratios (see~\autoref{tab:agent}).

\begin{table}[!htbp]
\centering
\caption{Ratio of feasible task and traversable edges per agent type}
\label{tab:agent}
\scalebox{0.9}{
\begin{tabular}{lrr}
\hline
\rowcolor[HTML]{EFEFEF} 
Agent Type  & \multicolumn{1}{l}{\cellcolor[HTML]{EFEFEF}Feasible Task Ratio} & \multicolumn{1}{l}{\cellcolor[HTML]{EFEFEF}Edge Traversable Ratio} \\ \hline
\rowcolor[HTML]{FFFFFF} 
Human       & 1    & 1    \\
\rowcolor[HTML]{EFEFEF} 
Wheelchair  & 0.48 & 0.71 \\
\rowcolor[HTML]{FFFFFF} 
Humanoid    & 0.22 & 0.43 \\
\rowcolor[HTML]{EFEFEF} 
Quadrupedal & 0.97 & 0.96 \\
\rowcolor[HTML]{FFFFFF} 
Sweeper     & 0.57 & 0.79 \\ \hline
\end{tabular}
}
\end{table}

% \subsection{Task Formulation}
% \label{subsec:task-formulation}

% \noindent\textbf{Input:} video $\mathcal{X}$ at $2$\,FPS (or a specified frame subset), graph $\mathcal{G}$, task $\tau$ with grounded $(s,g)$, embodiment $\mathbf{a}$, and an output-format prompt. 

% \noindent\textbf{Output:} a JSON object with:
% \begin{itemize}[leftmargin=1.5em,itemsep=0pt,topsep=2pt]
%     \item \texttt{feasible} $\in \{\texttt{true},\texttt{false}\}$;
%     \item \texttt{route}: list of node IDs $[v_0,\dots,v_m]$ (proposed path $s\!\rightarrow\!g$);
%     \item \texttt{reason} (optional if \texttt{feasible}=\texttt{false}): short rationale, ideally referencing the failure taxonomy.
% \end{itemize}

% \noindent\textbf{Example JSON:}
% \begin{verbatim}
% {
%   "feasible": false,
%   "route": ["entrance_foyer", "hallway_1", "kitchen_door"],
%   "reason": "Door to kitchen too narrow for wheelchair."
% }
% \end{verbatim}

\subsection{Metrics \& Scoring}
\label{subsec:metrics}

We evaluate model outputs against ground truth using four complementary metrics, reported both overall and per-embodiment. Let
$\mathcal{G}=(\mathcal{V},\mathcal{E})$ be the space graph, let
$P(v_{\mathrm{src}},v_{\mathrm{tgt}})$ denote a simple path in $\mathcal{G}$, and let
$E(\hat{P})\subseteq \mathcal{E}$ be the ordered edge set of a predicted path $\hat{P}$.

\paragraph{(1) Feasibility Classification (\textbf{Feas-F1}).}
Let $\hat{y}\in\{0,1\}$ be the model’s feasibility prediction and $y^\star$ the ground truth (Sec.~\ref{subsec:gt}). We report positive-class precision, recall, and $F_1$:

\paragraph{(2) Path Validity (\textbf{PV}).}
A predicted route $\hat{P}=[v_0,\ldots,v_m]$ is considered valid if  
(i) $v_i\!\in\!\mathcal{V}$ for all $i$,  
(ii) $(v_i,v_{i+1})\!\in\!\mathcal{E}$ for all $i$,  
(iii) $v_0\!=\!v_{\mathrm{src}}$ and $v_m\!=\!v_{\mathrm{tgt}}$, and  
(iv) $\hat{P}$ is a simple path (no node repetitions).  
We report:
\[
\mathrm{PV} = \mathbb{E}\big[\mathbb{I}(\hat{P} \in \mathcal{P}_{\text{simple}}(v_{\mathrm{src}}, v_{\mathrm{tgt}}))\big].
\]

\paragraph{(3) Route Traversability Accuracy (\textbf{RTA}).}
Conditioned on $\hat{y}=1$ and a valid $\hat{P}$, we evaluate the fraction of edges in $\hat{P}$ that are actually traversable under embodiment $\mathbf{a}$:
\[
\mathrm{RTA}(\hat{P},\mathbf{a})
\;=\;
\frac{\sum_{e\in E(\hat{P})} g_e^{(\mathbf{a})}}{|E(\hat{P})|},
\]
where $g_e^{(\mathbf{a})}\!\in\!\{0,1\}$ is the edge-level traversability label.  
All edges are assigned unit weight so each edge contributes equally.  
$\mathrm{RTA}=1$ indicates a fully valid route, while lower values quantify partial failures.  
This metric serves a role analogous to path-quality measures such as SPL \citep{anderson2018r2r}, but conditions on embodiment-specific traversability.

\paragraph{(4) Reasoning Validity (\textbf{RV}).}
For infeasible predictions ($\hat{y}=0$), the model must (i) return a proposed route $\hat{P}$ and (ii) provide a short rationale $\hat{\rho}$ explaining the failure. Let $\mathcal{R}^\star(\hat{P},\mathbf{a})$ be the set of annotated failure reasons, we define an LLM-as-judge function $J_{\text{LLM}}$ that returns $1$ if the model’s rationale semantically matches the annotation and $0$ otherwise. We also verified the reliability of this function by manually validating 300 randomly sampled VLM reasonings. The results show an 89\% alignment between human and LLM verdicts, indicating the precision and effectiveness of the LLM-as-judge method. We report RV as the mean of binary judgments:
\[
\mathrm{RV} \;=\; \mathbb{E}\!\left[J_{\text{LLM}}\!\big(\hat{\rho}, \mathcal{R}^\star(\hat{P},\mathbf{a})\big)\right].
\]

%\paragraph{Format Error Rate (\textbf{FER}).}
%We additionally flag JSON parse errors, missing required keys, or hallucinated node IDs ($\notin\mathcal{V}$). Any such issue marks the instance as a format error; $\mathrm{FER}$ is the fraction of errors over all instances.% By default, FER-gated instances are scored as invalid for PV/RTA and as incorrect for Feas-F1.

\paragraph{Composite CapNav Score.}
We report a composite score that aggregates the above aspects:
\[
\mathrm{CapNav} \;=\;
\lambda_c\, F_1 \;+\; \lambda_p\, \mathrm{PV} \;+\; \lambda_t\, \overline{\mathrm{RTA}} \;+\; \lambda_r\, \overline{\mathrm{RV}},
   \sum \lambda_\cdot = 1,
\]
where $\overline{\mathrm{RTA}}$ averages over positive predictions, and $\overline{\mathrm{RV}}$ averages over negative predictions. We report per-embodiment scores and a macro-average across embodiments. By default, we set all $\lambda$ as 0.25.

\paragraph{Reference Upper and Lower Bounds.}
To contextualize model performance, we establish both lower and upper reference bounds under the proposed evaluation metrics. As a lower bound, we simulate a random-walk policy on the navigation graph that ignores scene structure and agent capability constraints. This naive baseline achieves a CapNav score of 0.29. As an approximate upper bound, we evaluate human performance on the same task. We recruited four human participants, each completing 20 randomly sampled navigation tasks using the same input and output format as the models. Human CapNav scores range from 0.49 to 0.75 (\textit{AVG} = 0.61).

%% file: sec/4_experiments_new.tex
\input{tables/performance}
\section{Experiments}
Our experiments address three research questions:
\begin{enumerate}
    \item \textbf{RQ1:} How do state-of-the-art VLMs perform on \textsc{CapNav}?
    \item \textbf{RQ2:} How does performance vary across models, agent types, and inference settings?
    \item \textbf{RQ3:} What failure modes expose the current limitations of these models?
\end{enumerate}
To answer these questions, we evaluate 13 modern VLMs on \textsc{CapNav}. We report quantitative results stratified by model family, agent type, inference setting (frame budget and thinking mode), and challenge type. We further analyze common failure cases and consolidate them into a qualitative error taxonomy.

\subsection{Settings}
We evaluated 13 VLMs in November 2025, spanning popular proprietary and open-source models with strong multimodal capabilities, including the Gemini~2.5 family, GPT 4.1 and 5-Pro, Doubao-Seed, and Qwen3-VL. For models that expose a thinking option, we report results under both \emph{thinking} and \emph{non-thinking} configurations. We additionally include Spatial-MLLM~\cite{wu2025spatial} and Video-R1~\cite{feng2025video}, which explicitly target spatial reasoning---a core requirement of \textsc{CapNav}. Proprietary models are accessed via APIs, while open-source models are deployed on either a single NVIDIA A100 GPU (80GB) or a single A40 GPU (48GB).

At inference time, each model receives \textsc{CapNav} tasks as a triple $\langle \mathcal{S}, \tau, \mathbf{a} \rangle$ (Section~\ref{subsec:probstate}). Depending on model I/O constraints, $\mathcal{S}$ is provided either as a video clip or as a set of uniformly sampled frames. A practical limitation of current VLM interfaces is the maximum number of visual tokens/frames accepted per request. To study this constraint, we evaluate up to four input settings for each model (subject to interface limits): \textbf{16}, \textbf{32}, and \textbf{64} total frames per clip, and a \textbf{1 FPS} setting that samples the full video (avg.\ 163 frames/video).

\subsection{Performance}
Using the metrics in Section~\ref{subsec:metrics}, we compare model outputs against the ground-truth navigation graph to compute overall and per-agent performance. Table~\ref{tab:benchscore} summarizes performance metrics across all models under both \emph{thinking} and \emph{non-thinking} settings. We show the best metrics when multiple input frame settings are tested. 

All evaluated models outperform the random-walk baseline (CapNav = 29.35\%). The strongest performance is achieved by proprietary models (notably the Gemini-2.5 family and GPT-5-pro), which exceed the human average (CapNav = 60.59\%) but remain below the best human performance (CapNav = 74.77\%). Overall, results show a clear separation between top-tier closed-source systems and the remainder of the field.

Models explicitly designed for spatial reasoning (Spatial-MLLM and Video-R1-7B) perform substantially worse across all metrics, despite claims of strong spatial performance. This indicates that their spatial reasoning training schemes and architectural modifications are not yet sufficient to address the challenges posed by \textsc{CapNav}.

We also test the sensitivity of rankings to the CapNav score composition. In addition to equal weighting, we evaluate four alternative schemes that assign weight $0.5$ to one component and $1/6$ to each of the remaining components. Kendall's $\tau$ between rankings across these schemes averages $0.909$, suggesting that conclusions are largely robust to moderate weight changes.

\subsection{Failure-mode taxonomy}
To characterize systematic errors, we use Gemini-3 to mine recurring error themes from 1{,}000 randomly sampled inference outputs, and then manually refine these themes into a four-category taxonomy. We then visualize the VLM predicted results of 1500 QA pairs sampled under 10 model settings in 3D scenes to measure the prevalence and distribution of each error type.
\begin{itemize}
    \item \textbf{Path hallucination:} (N = 659 / 1500) the predicted path links non-adjacent nodes, or violates the specified start/end, indicating an incorrect reasoning of the scene connectivity.
    \item \textbf{Obstacle hallucination:} (N = 418) the model reports non-existent blocking obstacles (e.g., ``closed doors'' or severe clutter), suggesting unreliable spatial interpretation of visual evidence.
    \item \textbf{Dimension neglect:} (N = 191) the model overlooks geometric constraints that block the agent (e.g., narrow doors, tight clearances, cluttered passages), reflecting weak grounding between spatial extent and traversability.
    \item \textbf{Ability hallucination:} (N = 10) the output reasoning contradicts the provided agent profile. This failure is only observed in smaller open-sourced models (e.g., MiMo-VL 7B).
\end{itemize}

\subsection{Physical Obstacle Types}
\begin{figure}
    \centering
    \includegraphics[width=0.75\linewidth]{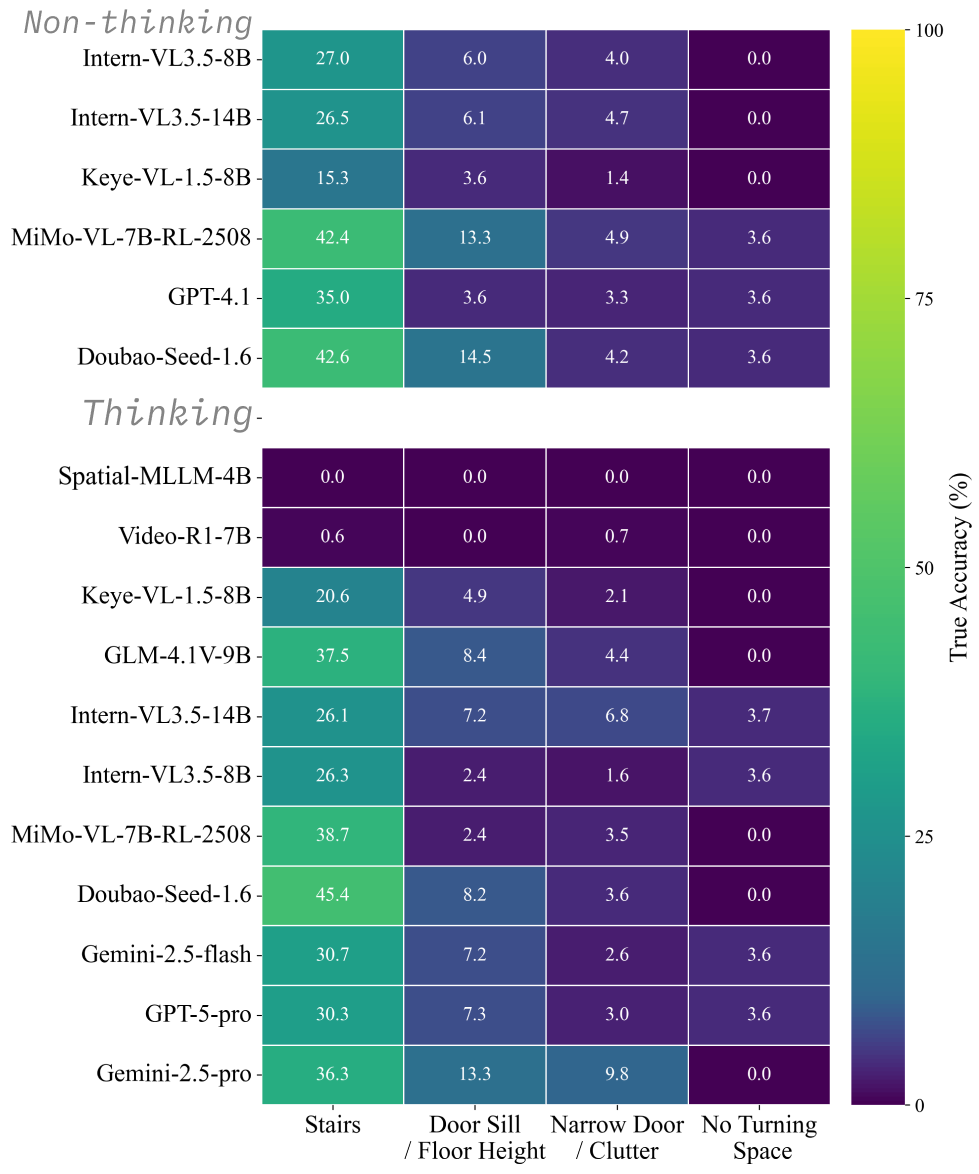}
    \caption{Model accuracy by obstacle type.}
    \label{fig:obstacle}
\end{figure}

We further stratify by major obstacle categories in the traversability ground truth. We observe four major types of ground-truth obstacles: \textbf{stairs} (N = 520 / 2365), which block humanoid robots, sweeping robots, and wheelchair users; \textbf{door sill/floor height differences} (N = 82), which stop wheelchair users and sweeping robots; \textbf{narrow pathways} (N = 438), usually caused by narrow doors or furniture clutter, which can restrict humanoid robots, and also wheelchair users in extreme cases; \textbf{lacking of turning spaces} (N = 28), which hinders wheelchair users and, in rare cases, quadrupedal robots. For the navigation tasks labeled as non-traversable due to one or more of these obstacles, we evaluate how often VLMs can both correctly predict infeasibility and provide appropriate reasoning. The results are visualized in \autoref{fig:obstacle}.

%% file: tables/performance.tex
\begin{table*}[ht!]
\centering
% \small
\scalebox{0.8}{
\begin{tabular}{lccccc|ccccc}
\toprule
\textbf{Model} & \textbf{Feas-F1} & \textbf{PV} & \textbf{RTA} & \textbf{RV} & \textbf{CapNav} &
\textbf{HUMAN} & \textbf{HUMANOID} & \textbf{QUAD} & \textbf{SWEEP} & \textbf{WHEEL} \\
\cmidrule(lr){2-6} \cmidrule(lr){7-11}

% ----------- Block 1: Non-thinking mode -----------
\multicolumn{11}{l}{\textbf{Non-thinking mode}} \\
\midrule
\cellcolor{blue!10} Intern-VL3.5-8B %{\hypersetup{citecolor=blue}\cite{zhu2025internvl3}}
\cite{zhu2025internvl3}
& 73.65 & 26.53 & 39.54 & 27.17 & 41.72 
& 53.89 & 35.61 & 51.92 & 51.36 & 49.43 \\

\cellcolor{blue!10} Intern-VL3.5-14B \cite{zhu2025internvl3} 
& 75.68 & 35.82 & 50.21 & 25.18 & 46.72
& 50.43 & 42.98 & 47.69 & 44.66 & 47.38 \\

\cellcolor{blue!10} Keye-VL-1.5-8B   \cite{yang2025kwai}
& 77.63 & 30.84 & 43.96 & 36.70 & 47.28
& 58.64 & \textbf{53.44} & 54.37 & \textbf{62.17} & 60.11 \\

\cellcolor{blue!10} MiMo-VL-7B-RL-2508  \cite{coreteam2025mimovltechnicalreport} 
& 57.31 & 43.61 & 60.08 & 29.35 & 47.59
& 46.61 & 38.96 & 45.38 & 41.46 & 50.85 \\

\cellcolor{blue!10} Qwen3-VL-8B-Instruct  \cite{qwen3technicalreport}  
& 78.92 & 43.39 & 53.56 & \textbf{47.89} & 55.94
& 54.20 & 38.18 & 50.91 & 57.50 & 53.21 \\

\cellcolor{green!10} GPT-4.1       
\cite{openai2025gpt41}
& 75.90 & 49.00 & 65.51 & 32.86 & 55.82
& \textbf{73.49} & 40.35 & 49.74 & 57.70 & 56.24 \\

\cellcolor{green!10} Doubao-Seed-1.6 \cite{bytedance2025seed16}
& \textbf{80.26} & \textbf{60.00} & \textbf{71.92} & 35.47 & \textbf{61.91}
& 65.18 & 47.23 & \textbf{59.49} & 60.65 & \textbf{60.82} \\
\midrule

% ----------- Block 2: Thinking mode -----------
\multicolumn{11}{l}{\textbf{Thinking mode}} \\
\midrule

\cellcolor{orange!10} Spatial-MLLM-4B         
\cite{wu2025spatial}
& 75.27 & 5.04 & 10.16 & - & 30.15
& 35.87 & 14.86 & 29.48 & 25.78 & 22.69 \\

\cellcolor{orange!10} Video-R1-7B    \cite{feng2025video}
& 74.57 & 25.50 & 44.66 & 4.82 & 37.39
& 43.62 & 26.35 & 41.66 & 32.17 & 31.82 \\

\cellcolor{blue!10} Keye-VL-1.5-8B   \cite{yang2025kwai}      
& 73.07 & 41.61 & 48.54 & 38.67 & 50.47
& 51.05 & 30.99 & 47.59 & 46.52 & 45.98 \\

\cellcolor{blue!10} GLM-4.1V-9B         \cite{vteam2025glm45vglm41vthinkingversatilemultimodal}
& 73.81 & 40.12 & 55.43 & 33.12 & 50.62
& 50.90 & 33.19 & 46.01 & 45.47 & 42.46 \\

\cellcolor{blue!10} Intern-VL3.5-14B \cite{zhu2025internvl3}    
& 78.48 & 42.96 & 57.90 & 24.97 & 51.08
& 50.21 & 39.93 & 51.13 & 46.27 & 47.67 \\

\cellcolor{blue!10}{Intern-VL3.5-8B \cite{zhu2025internvl3}}
 
& 79.55 & 43.74 & 55.10 & 36.50 & 53.72
& 51.23 & 35.34 & 51.61 & 53.74 & 50.05 \\

\cellcolor{blue!10} MiMo-VL-7B-RL-2508    \cite{coreteam2025mimovltechnicalreport}
& 65.10 & 59.62 & 65.00 & 35.33 & 56.26
& 46.61 & 38.96 & 45.38 & 41.46 & 50.85 \\

\cellcolor{green!10} Doubao-Seed-1.6 \cite{bytedance2025seed16} 
& 76.16 & 61.94 & 71.93 & \textbf{38.44} & 62.12
& 60.31 & 46.20 & 58.05 & 61.02 & 63.35 \\

\cellcolor{green!10} Gemini-2.5-flash    
\cite{comanici2025gemini}
& 84.16 & 65.12 & 68.96 & 38.04 & 64.07
& 79.95 & 40.21 & 59.15 & 63.07 & 59.46 \\

\cellcolor{green!10} GPT-5-pro       \cite{openai2025gpt5}
& \textbf{86.87} & 67.90 & 75.89 & 34.81 & 66.37
& 82.85 & 46.95 & \textbf{61.74} & 70.58 & 64.78 \\

\cellcolor{green!10} Gemini-2.5-pro  \cite{comanici2025gemini}
& 84.30 & \textbf{73.00} & \textbf{79.15} & 32.29 & \textbf{67.18}
& \textbf{85.96} & \textbf{54.47} & 61.21 & \textbf{70.87} & \textbf{65.51} \\

\bottomrule
\end{tabular}
}
\caption{
CapNav performance across 13 VLMs. Each number indicate the best performance under all tested frame settings.
Left block reports each task metrics (Feas-F1, PV, RTA, RV, CapNav). 
Right block reports per-agent composite scores for the five agent types: adult with no motor disabilities, humanoid robot, quadrupedal, sweeping robots, and wheelchair users. 
Bold number indicate best per block. Blue indicate open-sourced model; Green indicate proprietary; Orange indicate spatial reasoning models.
}
\label{tab:benchscore}
\end{table*}

%% file: sec/5_findings_polished.tex
\section{Findings}
Drawing on both quantitative results and qualitative inspection of model outputs, we distill three primary findings. Collectively, they indicate that current VLM-based navigation remains brittle under capability constraints, visually demanding inputs, and geometry-sensitive traversability judgments. We also summarize actionable guidance for practitioners and outline concrete directions for future work.

\subsection{Capability constraints induce systematic performance degradation}

Across all tested models, performance is highest under the baseline embodiment without mobility limitations (adults without mobility constraints; mean \textsc{CapNav} score $=57.83\%$). As constraints tighten, performance degrades. The largest gap appears for the \textsc{HUMANOID} embodiment, which diverges most from the \textsc{HUMAN} setting due to (i) an inability to climb stairs and (ii) a minimum pathway clearance requirement of $0.9\,\mathrm{m}$; this configuration attains the lowest average score (mean $=39.12\%$).

This pattern highlights systematic deficits and imbalances in VLM-based navigation under mobility constraints. Strong performance in human-like settings does not translate to capability-constrained scenarios; hence, targeted evaluation is essential before VLM deployment in safety-critical navigation.

The five agent types supported in \textsc{CapNav} are chosen to represent common human and robotic mobility profiles. As such, our results generalize to a wide range of practical embodiments. Practitioners may approximate expected model performance for their own systems by aligning their agent's capabilities with the closest of the five provided profiles. For embodiments that differ substantially, \textsc{CapNav}'s open-sourced annotation interface and data curation pipeline allow users to define new agent profiles and extend the benchmark. This facilitates capability-aware evaluation for diverse hardware platforms and emerging robotic systems.

\subsection{Increasing visual budget helps, but exposes a visual bottleneck}

Comparing matched inference settings with and without \emph{thinking}, we find consistent improvements when \emph{thinking} is enabled. Across 9 matched pairs, \emph{thinking} configurations outperform their non-\emph{thinking} counterparts by an average of $\Delta$\textsc{CapNav} $=6.87\%$. This gain is accompanied by a substantial compute cost: in our measurements, mean inference time increases by approximately $8\times$ (from $14.94\,\mathrm{s}$ to $123.94\,\mathrm{s}$).

\begin{figure}[t]
    \centering
    \includegraphics[width=\linewidth]{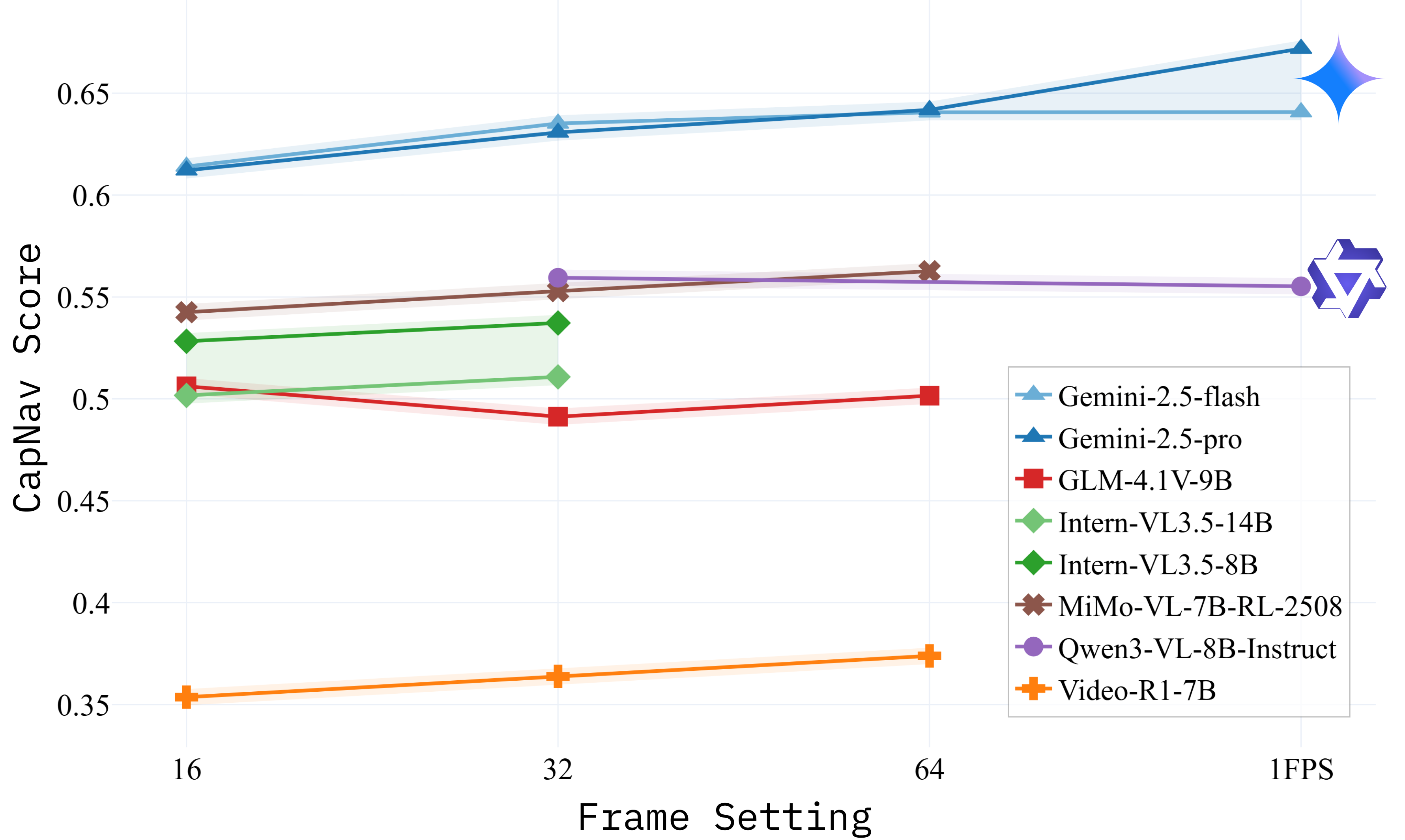}
    \caption{Performance across frame-budget settings. While higher frame counts can improve results, the gains are model-dependent and may saturate, consistent with a visual integration bottleneck.}
    \label{fig:frame_setting_performance}
\end{figure}

Across the \textbf{16}/\textbf{32}/\textbf{64}/\textbf{1\,FPS} frame settings, latency is comparatively stable, but higher frame counts increase GPU VRAM and API costs. Performance generally improves with more frames; however, the marginal benefit is uneven, tending to be larger for stronger models (notably Gemini) and smaller or negligible for weaker models (see \autoref{fig:frame_setting_performance}). This pattern reveals a \emph{visual bottleneck}: additional evidence only helps if the model can reliably integrate it.

Qualitative inspection suggests two explanations:
(i) For most open-sourced models, 64 frames is already a very heavy input but it can still remain sparse in large or complex scenes, requiring robust cross-frame spatial integration and viewpoint alignment that weaker models often lack.
(ii) Additional viewpoints can increase \emph{obstacle hallucinations} in some models: salient but non-blocking cues may be misinterpreted as traversal barriers, echoing prior observations about limitations in nonlocal reasoning for VLMs \cite{berman2025vlms}.

These results imply a cost--benefit frontier rather than a monotonic ``more is better'' rule. For \textsc{CapNav}-like VLN tasks, increasing inference budget can improve performance, but the return depends on model capacity and may trade off against latency, memory, cost, and even error modes. 

\subsection{Dimension neglect: models under-detect geometry-driven constraints}

From our obstacle type analysis in \autoref{fig:obstacle}, all tested models perform markedly better on obstacles with salient visual signatures (e.g., stairs, door sills) than on constraints requiring implicit metric estimation (e.g., narrow clearances, turning radii). These failures indicate a recurring form of \emph{dimension neglect}: traversability decisions that hinge on estimating size, clearance, or maneuvering space from images remain challenging for all tested VLMs.

% Promising routes to address dimension neglect include:
% \begin{itemize}
%     \item \textbf{Explicit geometric scaffolding:} inject depth/metric priors (e.g., monocular depth, coarse 3D reconstructions, or calibrated scale cues) and compute clearance/turn feasibility via lightweight geometric modules rather than relying on implicit reasoning alone.
%     \item \textbf{Capability-aware representation learning:} train models to represent traversability as a function of both scene geometry and embodiment parameters (clearance threshold, turning radius), encouraging systematic generalization across profiles.
%     \item \textbf{Targeted supervision for metric judgments:} add auxiliary objectives that reward correct identification of dimension-critical constraints and penalize ``near-miss'' errors (e.g., confusing a visually narrow passage with a passable one).
% \end{itemize}

%\subsection{A pilot fine-tuning study shows partial learnability but introduces new failure modes}

To probe whether the above deficits are learnable, we ran a pilot fine-tuning experiment. We augment the CapNav dataset on the navigation graph to expand the task count from 2K to 13K, and LoRA fine-tune on Qwen3-VL-8B-Instruct. After one epoch of training, test score improves from $45.26\%$ to $55.18\%$, and the finetuned model generates more dimension-aware reasoning. However, reasoning validity decreases ($0.30 \rightarrow 0.25$), and we observe frequent false positives on narrow passages (i.e., incorrectly flagging feasible routes as blocked).

These outcomes suggest that fine-tuning can reduce dimension neglect but does not fully resolve geometric precision, and may shift the error profile toward overly aggressive obstacle judgments. This highlights the need to jointly optimize for \emph{obstacle awareness} and \emph{dimension reasoning}, rather than treating score improvements alone as sufficient.

Building on this pilot, we plan to explore RL-based tuning with task-level rewards under embodiment constraints. We also aim to experiment with spatial-information injection (depth estimates, topological priors, or map-like intermediate representations) to improve spatial dimension judgments.

%% file: sec/6_conclusion.tex
% \section{Conclusion}

% We present \textsc{CapNav}, a benchmark that systematically evaluates capability-conditioned navigation across diverse embodiments, realistic mobility obstacles, and graph-structured indoor environments. By assessing 13 state-of-the-art VLMs, we show that while modern VLMs exhibit strong navigation performance in human-like settings, they fail to generalize when physical capabilities diverge. Performance degrades for more constrained agents and obstacle recognitions that requires spatial dimension reasoning, and under limited visual input budgets—revealing gaps that current VLMs can only partially mitigate. These findings underscore the risks of applying general-purpose VLMs as drop-in navigation planners without capability-aware evaluation, and highlight the need for models that reason robustly about physical dimensions, environmental constraints, and embodiment-specific affordances. By releasing the benchmark, dataset, and annotation interfaces, \textsc{CapNav} establishes a foundation for developing and evaluating next-generation VLMs that support safer, more inclusive, and capability-aware navigation in complex real-world spaces.

\section{Conclusion}

We present \textsc{CapNav}, a benchmark that systematically evaluates capability-conditioned navigation across diverse embodiments, realistic mobility obstacles, and graph-structured indoor environments. By evaluating 13 state-of-the-art VLMs, we observe three consistent patterns. First, performance degrades systematically as mobility constraints tighten, indicating that strong results in unconstrained human-like settings do not transfer reliably to other embodiments. Second, increasing the inference budget (more frames or ``thinking'' mode) yields inconsistent gains and incurs computational cost, exposing a visual integration bottleneck. Third, models exhibit dimension neglect: they handle salient obstacles better but struggle with dimension reasoning (\textit{e.g.}, narrow clearances and turning radii). A pilot fine-tuning study suggests that these deficits are partially learnable, but fine-tuning may introduce false positives, highlighting the need to jointly optimize the spatial reasoning reliability in task formulation and spatial dimension. We release both the \textsc{CapNav} dataset and annotation tool.

%% file: sec/X_suppl.tex
\clearpage
\setcounter{page}{1}
\maketitlesupplementary

%\section{Rationale}
%\label{sec:rationale}
% 
%Having the supplementary compiled together with the main paper means that:
% 
%\begin{itemize}
%\item The supplementary can back-reference sections of the main paper, for example, we can refer to \cref{sec:intro};
%\item The main paper can forward reference sub-sections within the supplementary explicitly (e.g. referring to a particular experiment); 
%\item When submitted to arXiv, the supplementary will already included at the end of the paper.
%\end{itemize}
% 
%To split the supplementary pages from the main paper, you can use \href{https://support.apple.com/en-ca/guide/preview/prvw11793/mac#:~:text=Delete%20a%20page%20from%20a,or%20choose%20Edit%20%3E%20Delete).}{Preview (on macOS)}, \href{https://www.adobe.com/acrobat/how-to/delete-pages-from-pdf.html#:~:text=Choose%20%E2%80%9CTools%E2%80%9D%20%3E%20%E2%80%9COrganize,or%20pages%20from%20the%20file.}{Adobe Acrobat} (on all OSs), as well as \href{https://superuser.com/questions/517986/is-it-possible-to-delete-some-pages-of-a-pdf-document}{command line tools}.

\section*{Overview of Supplementary Material}

This supplementary material provides examples of VLM input and output in the dataset generation and benchmark evaluation process. We include the following components: 
(1) the exact prompts and input representations provided to VLMs, 
(2) examples of ground truth

\section*{A. Task Generation}
\label{sec:A}
\subsection*{A.1 Prompt Template}
% We use a unified text prompt to instruct a vision-language model to
% generate capability-conditioned navigation questions for each scene. The template
% contains three components: (1) the instruction, (2) the materials
% provided to the model (video, guideline PDF, and scene nodes, specified in Sec A.2-4), and
% (3) variable fields populated for each scene.

We use a unified prompt template to instruct a VLM to
generate navigation tasks for each scene. The template
contains three components: (1) an overall instruction, (2) file attachments, \ie a scene video and a guideline document that contains more detailed instructions (Sec A.2-3), and
(3) a list of spatial nodes in the scene (Sec A.4).

An example of generated tasks can be found in \path{taskGeneration/HM3D00025_tasks.json}.

\begin{tcolorbox}[
    colback=gray!3,
    colframe=gray!40,
    left=2mm,
    right=2mm,
    top=1.5mm,
    bottom=1.5mm,
    arc=1mm,
    boxrule=0.4pt,
    breakable
]

% \textbf{Instruction:}\\[0.3em]
% You are an expert benchmark question designer for the 
% Capability-Conditioned Navigation Benchmark.  
% You are provided with three materials:
% \begin{enumerate}[leftmargin=1.2em, itemsep=1pt, topsep=2pt]
%     \item A video showing the indoor environment.
%     \item A guideline PDF defining how to create navigation questions.
%     \item A list of valid scene nodes.
% \end{enumerate}

% \vspace{0.5em}

% \textbf{Requirements:}\\[0.3em]
% Please strictly follow the guideline and generate realistic, visually
% grounded, route-based navigation questions. Each question must:
% \begin{itemize}[leftmargin=1.2em, itemsep=1pt, topsep=2pt]
%     \item Include a start and end node with detailed in-room descriptions.
%     \item Follow the schema structure provided by the system.
%     \item Output a valid JSON array (no extra commentary).
%     \item Ensure the total number of generated questions is $\ge$ \{min\_questions\}.
% \end{itemize}

% \vspace{0.5em}

% \textbf{Node List (variable):}\\[0.3em]
% \{node\_list\_text\}

\textbf{Instruction:}\\[0.3em]
You are an expert benchmark question designer for the 
Capability-Conditioned Navigation Benchmark.  
You are provided with three materials:
\begin{enumerate}[leftmargin=1.2em, itemsep=1pt, topsep=2pt]
    \item A video showing the indoor environment.
    \item A guideline PDF defining how to create navigation questions.
    \item A list of valid scene nodes.
\end{enumerate}

\vspace{0.5em}

Requirements:\\[0.3em]
Please strictly follow the guideline and generate realistic, visually
grounded, route-based navigation questions. Each question must:
\begin{itemize}[leftmargin=1.2em, itemsep=1pt, topsep=2pt]
    \item Include a start and end node with detailed in-room descriptions.
    \item Follow the schema structure provided by the system.
    \item Output a valid JSON array (no extra commentary).
    \item Ensure the total number of generated questions is $\ge$ \{min\_questions\}.
\end{itemize}

\vspace{0.5em}

\textbf{External files (video, pdf)}
\vspace{0.5em}

\textbf{Node List:}\\[0.3em]
\{node\_list\_text\}

\end{tcolorbox}

% ===============================
% A.2 Video Input Format
% ===============================

\subsection*{A.2 Video Input}

\begin{figure}[h]
\centering
\includegraphics[width=\linewidth]{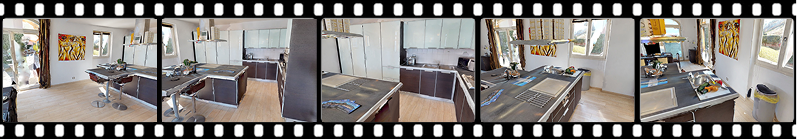}
\caption{Representative frames from the input
video \texttt{HM3D00025.mp4}. The VLM receives the complete MP4 video for navigation task generation.}
\label{fig:video-frames}
\end{figure}

When generating navigation tasks, the VLM receives a touring video of the space.
% In Fig.\ref{fig:video-frames}, we show a short excerpt of the video in frame form and provide a single stitched image that visualizes representative frames from the video.
In Fig.\ref{fig:video-frames}, we show representative frames of an example video. 
The full MP4 file is included in the supplementary ZIP, see \path{taskGeneration/HM3D00025.mp4}.

% \begin{tcolorbox}[colback=gray!3, colframe=gray!40, breakable]
% \small
% \textbf{Video Frames for Scene HM3D00025 (excerpt):}
% \begin{PromptJSON}
% frame_0001.jpg
% frame_0002.jpg
% ...
% frame_0254.jpg
% frame_0255.jpg
% (Full MP4 video provided in supplementary ZIP)
% \end{PromptJSON}

% \end{tcolorbox}

% ===============================
% A.3 PDF Guideline
% ===============================

\subsection*{A.3 Guideline Document}
% During question generation, the model receives the full guideline
% document that specifies how to compose capability-conditioned navigation
% questions. The complete PDF is included in the supplementary ZIP
% (\texttt{Generate\_QA\_Guidelines.pdf}). Below we show an excerpt of the
% text provided to the model during inference.
We provide a carefully designed guideline document to the VLM to specify the exact type and format of the navigation tasks to be generated. We show an excerpt of the document below. The complete guideline is included in the supplementary ZIP 
% (\texttt{questionGeneration/Generate\_QA\_Guidelines.pdf}).
(\path{taskGeneration/Generate_Tasks_Guidelines.pdf}).

\begin{tcolorbox}[colback=gray!3, colframe=gray!40, breakable]
\small
\textbf{Guideline (excerpt):}\\[-0.3em]

\begin{PromptTEXT}
Capability-conditioned navigability

1. Goal
This guideline defines how to generate capability-neutral, route-based navigation questions for the Capability-Conditioned Navigation Benchmark. You are an AI model that generates route-based navigation questions from an indoor walkthrough video.
Your task is to create realistic and scene-consistent navigation tasks that test how [Agent] can move and act within the visible environment.

Each question must:
1. Be fully grounded in the video (only describe objects, rooms, and connections that actually appear).
2. Use the provided list of scene nodes as the only valid room-level landmarks.
...

\end{PromptTEXT}
Note: Full guideline PDF in the supplementary ZIP.
\end{tcolorbox}

\subsection*{A.4 Node List}

To ground the sub-spaces in a scene, the VLM also receives a list of node IDs paired with textual descriptions. We instruct the VLM to reference these nodes when generating navigation tasks. Below we show a few nodes from the scene \texttt{HM3D00025}, taken directly from the scene
graph. The full JSON file is included in the supplementary ZIP
(\path{groundTruth/HM3D00025-graph.json}).

\begin{tcolorbox}[colback=gray!3, colframe=gray!40, breakable]
\small
\textbf{Node List:}

\begin{PromptJSON}
{
node_117: Master bedroom foyer
node_118: Master bedroom bathroom
node_119: Master bedroom home office
node_120: Top floor stair landing and hallway
...
node_131: First floor bedroom with twin beds
node_132: First floor bathroom
}

\end{PromptJSON}
Note: We show a subset of the nodes here. The model receives the full
node list during task generation.
\end{tcolorbox}

\section*{B. Benchmark Evaluation}
% Querying VLMs
When evaluating on the CapNav benchmark, we query VLMs to answer the 
capability-conditioned navigation questions. 
Each query provides the following inputs:

\begin{enumerate}[leftmargin=1.5em, itemsep=1pt, topsep=3pt]
    \item a specific indoor scene shown as a tour video (see \autoref{fig:video-frames}),
    \item all nodes of the scene's navigation graph,
    \item one navigation question,
    \item one agent profile describing the mobility capabilities.
    
\end{enumerate}

The model must use both the video and the text inputs to determine whether the agent can complete the navigation task. Below we show a prompt example for scene \texttt{HM3D00025} and agent \textsc{Humanoid}. The original prompt file can also be seen in \path{benchmarkEvaluation/HM3D00025_q01_HUMANOID.txt}

\subsection*{B.1 Example Prompt (HM3D00025, HUMANOID)}

\begin{tcolorbox}[
    colback=gray!3,
    colframe=gray!40,
    left=2mm,
    right=2mm,
    top=1.5mm,
    bottom=1.5mm,
    arc=1mm,
    boxrule=0.4pt,
    breakable
]

\textbf{Instruction:}\\[0.3em]
You are an expert visual reasoning agent for indoor navigation tasks. You will receive four materials:
\begin{enumerate}[leftmargin=1.2em, itemsep=1pt, topsep=2pt]
    \item A video showing the indoor environment.
    \item A single navigation question asking whether the agent can move from one area (node) to another.
    \item The agent's physical and capability profile.
    \item The list of all nodes in this environment, with their textual descriptions (e.g., room type, furniture, width of passages).
\end{enumerate}

Your goal is to determine whether the agent can successfully reach the destination area from the starting area,
based on both the video and the textual descriptions. You must reason step by step about spatial constraints,
obstacles, connectivity, and the agent’s mobility limitations.

\vspace{0.6em}

\textbf{Input:}
\\

\textbf{1.Navigation Question:}\\[0.3em]
Can [Agent] move from the dark sectional sofa area in the main living space to the white L-shaped sofa area in the first floor living room?

\vspace{0.6em}

\textbf{2.Agent Profile (HUMANOID):}\\[-1.6em]
\begin{PromptTEXT}
Agent name: HUMANOID  
Body shape: box  
Height (m): 1.5  
Width (m): 0.9  
...
Can operate elevator: True  
Can open the door: True  
Description: Boston Dynamics Atlas humanoid robot approximately the size of a human, capable of obstacle crossing up to 0.4 m, for a single obstacle. However, it cannot go up stairs.
\end{PromptTEXT}

\vspace{0.6em}

\textbf{3.Scene Node List:}\\[-1.6em]
\begin{PromptTEXT}
node_117 - Master bedroom foyer
node_118 - Master bedroom bathroom
node_119 - Master bedroom home office
...
node_131 - First floor bedroom with twin beds
node_132 - First floor bathroom
\end{PromptTEXT}

\vspace{0.6em}

\textbf{4.Video}
\\
(You can observe the video for spatial layout and obstacles.)
\vspace{0.6em}

\textbf{Task:}\\[0.3em]
Your goal is to determine whether the agent can \textbf{navigate} from the start area to the goal area.  
\\
Focus exclusively on \textbf{movement feasibility}, considering physical dimensions, obstacle heights, and connection constraints.\\
You must \textbf{not stop after a single failed route attempt}. If one possible route is blocked (e.g., by stairs or narrow spaces),
you must \textbf{actively consider all other possible paths} between the start and goal nodes in the scene graph.\\[0.3em]

Follow these principles:
\begin{enumerate}[leftmargin=1.2em, itemsep=1pt, topsep=2pt]
    \item Explore \textbf{all possible routes} through the scene graph before deciding the task is impossible. That means, try \textbf{multiple alternative routes} using all visible connections in the scene graph until you are confident that \textbf{no feasible route} exists.
    \item Account for the agent’s capabilities (e.g., door opening, elevator operation, stair traversal) when evaluating possible paths.
    \item When multiple feasible paths exist, select the \textbf{most direct and realistic one} given the agent's capabilities.
    \item If no route works, specify which \textbf{edge or physical barrier} prevents traversal and explain why.
    \item If a feasible route exists, specify the \textbf{sequence of nodes} representing the navigable path.
\end{enumerate}
\vspace{0.6em}

Important: \\
If you initially find the route impossible, \textbf{re-examine the scene graph} and attempt at least two distinct alternative paths before concluding "no".\\
Your reasoning should reflect persistent exploration: do not assume failure after one obstacle; explore until all logical alternatives are ruled out.\\
You do not need to consider unrelated interactions (e.g., turning on lights, using computers, or touching furniture).
\vspace{0.6em}

Please decide for each given question whether the agent can complete the navigation task.  \\
If yes, provide a \textbf{feasible path} through the relevant nodes.  \\
If no, specify the \textbf{edge (two connected nodes)} that blocks traversal and give a concise \textbf{reason} (e.g., too narrow passage, stairs, or closed door).  \\
Your reasoning should always consider the agent’s physical capabilities mentioned in the Agent profile(e.g., wheelchair cannot climb stairs, sweeper robot cannot open doors).

\vspace{0.6em}
Return your answer in the required structured JSON format below.

\vspace{0.6em}

\textbf{Output Format (JSON only):}\\[-0.3em]
\begin{PromptJSON}
[
    ...
    {
        "question": "...",
        "agent": "HUMANOID",
        "result": {
            "answer": "no",
            "path": ["node_12", "node_14", "node_15"],
            "reason": "Too narrow passage between the sofa and wall"
        }
    }
]
\end{PromptJSON}

Return only the JSON array, no explanation, commentary, or markdown formatting.

\end{tcolorbox}

\section*{C. Reasoning Evaluation}

CapNav deploys an LLM-as-judge method to evaluate whether a navigability explanation from a VLM's output is consistent with the
ground-truth annotations. We provide the LLM judge with (1) the VLM-generated
reasoning for why a path is non-traversable and (2) the full ground-truth traversability record for the
path. The LLM will determine whether the explanation correctly
captures the underlying failure conditions.

Below we briefly present the prompts and data formats. For a full example, please check \path{reasoningEvaluation/HM3D00025_q01_HUMANOID.txt}
\subsection*{C.1 Prompt Template}

\begin{tcolorbox}[colback=gray!3, colframe=gray!40, breakable]
\small
\textbf{Instruction (Reasoning Evaluation):}

\begin{PromptTEXT}
You are evaluating whether a navigation system's reasoning is correct.

You are given:
1. A system-generated explanation of why a path is not traversable.
2. The ground-truth traversability data for all edges along the path.

Your task:
- Determine if the system's reasoning correctly identifies or aligns with the actual traversability issues.
- Exact wording is not required; conceptual correctness is sufficient.
- Grant partial credit if the reasoning identifies some but not all issues.

Respond in JSON format with:
{
   "correct": true/false,
   "explanation": "Brief explanation of why the reasoning is correct or incorrect"
}
\end{PromptTEXT}
\end{tcolorbox}

\subsection*{C.2 Example Input (Excerpt)}

\begin{tcolorbox}[colback=gray!3, colframe=gray!40, breakable]
\small
\textbf{Example Input (excerpt):}

\begin{PromptTEXT}
System reasoning:
"The agent cannot traverse stairs, and the only path between the
Basement bar space (node_109) and the Laundry room (node_105)
requires going up the Basement stairs (node_114 or node_115),
which the agent cannot do."

Ground-truth traversability (excerpt):
{
  "total_edges": 6,
  "traversable": 4,
  "non_traversable": 2,
  "edges": [
    {
      "from": "node_109",
      "to": "node_108",
      "from_name": "Basement bar space",
      "to_name": "Pool and media room",
      "exists": true,
      "traversable": true,
      "note": "Edge from \"Basement bar space\" to \"Pool and media room\"",
      "ground_truth": {
        "exists": true,
        "traversable": true,
        "note": "Edge from \"Basement bar space\" to \"Pool and media room\""
      }
    },
    ...
  ]
}
\end{PromptTEXT}

\end{tcolorbox}